\documentclass[journal]{IEEEtran}
\usepackage[pdftex]{graphicx}
\usepackage{subfigure}
\usepackage{amsmath,amssymb,amsfonts}
\usepackage{color,xcolor}
\usepackage{blindtext}
\usepackage{algorithm}
\usepackage{algorithmic}
\usepackage{cite}
\usepackage{amsthm}
\usepackage[normalem]{ulem}
\usepackage{url}
\usepackage{comment}
\usepackage{multirow}
\usepackage{diagbox}
\usepackage{indentfirst}
\usepackage{ulem}
\usepackage{booktabs}
\usepackage{array}
\bstctlcite{IEEEexample:BSTcontrol}

\newcounter{ToDo}
\newcounter{gaocomm}
\newcounter{wangcomm}
\newcounter{Note}
\definecolor{blue-violet}{rgb}{0.54, 0.17, 0.89}
\definecolor{mygreen}{rgb}{0.0, 0.5, 0.0}
\definecolor{awesome}{rgb}{1.0, 0.13, 0.32}
\definecolor{bostonuniversityred}{rgb}{1.0, 0.0, 0.0}

\newcommand{\GaoC}[1]{\textcolor{blue-violet}{\stepcounter{gaocomm}{\bf [Junbin's comment \arabic{gaocomm}: #1]}\;}}

\title{CaEGCN: Cross-Attention Fusion based Enhanced Graph Convolutional Network for Clustering}

\begin{document}
\author{Guangyu~Huo, 
        Yong~Zhang,
        Junbin~Gao,
        Boyue~Wang,
        Yongli~Hu, and
        Baocai~Yin.
\IEEEcompsocitemizethanks{
\IEEEcompsocthanksitem Corresponding author: Boyue Wang.
\IEEEcompsocthanksitem Guangyu Huo, Yong Zhang, Boyue Wang, Yongli Hu and Baocai Yin are with Beijing Key Laboratory of Multimedia and Intelligent Software Technology, Beijing Artificial Intelligence Institute, Faculty of Information Technology, Beijing University of Technology, Beijing 100124, China. E-mail: gyhuo@emails.bjut.edu.cn, \{zhangyong2010,wby,huyongli,ybc\}@bjut.edu.cn.
\IEEEcompsocthanksitem Junbin Gao is with the Discipline of Business Analytics, The University of Sydney Business School, The University of Sydney, NSW 2006, Australia. E-mail: junbin.gao@sydney.edu.au
}
}

\markboth{IEEE Transactions on XX,~Vol.~XX, No.~X, January~2020}%
{Wang \MakeLowercase{\textit{et al.}}: Kernelized LRR on Grassmann }

\hyphenation{op-tical net-works semi-conduc-tor}
\maketitle

\begin{abstract}
With the powerful learning ability of deep convolutional networks, deep clustering methods can extract the most discriminative information from individual data and produce more satisfactory clustering results. However,  existing deep clustering methods usually ignore the relationship between the data. Fortunately, the graph convolutional network can handle such relationship, opening up a new research direction for deep clustering. In this paper, we propose a cross-attention based deep clustering framework, named Cross-Attention Fusion based Enhanced Graph Convolutional Network (CaEGCN), which contains four main modules: the cross-attention fusion module which innovatively  concatenates the Content Auto-encoder module (CAE) relating to the individual data and Graph Convolutional Auto-encoder module (GAE) relating to the relationship between the data in a layer-by-layer manner, and the self-supervised model that highlights the discriminative information for clustering tasks. While the cross-attention fusion module fuses two kinds of heterogeneous representation, the CAE module supplements the content information for the GAE module, which avoids the over-smoothing problem of GCN. In the GAE module, two novel loss functions are proposed that reconstruct the content and relationship between the data, respectively.
Finally, the self-supervised module constrains the distributions of the middle layer representations of CAE and GAE to be consistent. Experimental results on different types of datasets prove the superiority and robustness of the proposed CaEGCN.
\end{abstract}

\IEEEpeerreviewmaketitle

\section*{Supplementary materials}
The supplementary code is available at \url{https://github.com/huogy/CaEGCN}.

\section{Introduction}
\IEEEPARstart{C}lustering is an essential topic in data mining area, which divides a collection of objects into multiple clusters of similar objects.
Inspired by the powerful feature extraction capability of deep convolutional network, many deep learning based clustering methods have been proposed in recent years, demonstrating much significant progress in clustering research \cite{hinton2006reducing,xie2016unsupervised,ji2017deep,guo2017improved}. The two-step spectral clustering method is usually employed here: A `good' data representation or similarity matrix which is learned  from these deep learning methods can be pipelined to the downstream models/algorithms such as K-means \cite{macqueen1967some} or Normalized Cut \cite{tabatabaei2012ganc} to obtain the final clustering result.


However, existing deep clustering methods only focus on the data content, and usually ignore the relationship between the data, i.e., the structural information. 
With the development of the data collection and analysis technologies, people not only collect the data but also obtain or build the relationship between the data in the form of graphs, such as social networks \cite{girvan2002community}, biochemical structure networks \cite{velickovic2018graph} and railway networks \cite{zhang2020multi}.
These graphs can help people make better data-driven decisions.
Therefore, how to embed the relationship between the data into deep clustering becomes a thorny problem.


Furthermore, based on these raw graphs, one wishes to mine the latent relationship between the data effectively.
As we know, the edges in a graph represent the explicit relationship, which is also regarded as the first-order structural relationship.
Many graph embedding methods exploit such relationship, including DeepWalk \cite{perozzi2014deepwalk}, node2vec \cite{grover2016node2vec}, and LINE \cite{tang2015line}.
But, the data relationship in the real-world is complicated.
There still exist many implicit and complicated relationships.
For example, two nodes may  not be directly connected in a graph, while they have many identical neighbors. However it is natural belief that these two nodes have a high-order structural relationship.

In order to improve the clustering effect of deep clustering methods, utilizing the high-order relationships is necessary.
As an important approach in deep learning methods, Graph Convolutional Network (GCN) \cite{defferrard2016convolutional,bruna2014spectral,velickovic2018graph} can mine such potential high-order relationship between the data. GCN transfers the graph structured data to a low-dimensional, compact, and continuous feature space. While GCN simultaneously has very successful application in encoding and exploring graph structure and node content, it seems little attention has been given to applying GCNs to deep clustering tasks.


For the purpose of clustering, we can naturally construct an auto-encoder module based on a GCN, the so-called GAE module. GCN can lead the signals to be smoother, which is the inherent advantage of GCN.  However, such signal smoothing operation makes the signals more similar, losing the diversity of signals. It has been proven that GCN is prone to over-smoothing when the number of layers becomes large \cite{kipf2017semi}, which results in a poor performance in related tasks. 
So, GCN cannot be stacked as deeply as the CNN model in visual tasks.

To overcome this drawback of GCN, we introduce a common auto-encoder network to supplement the data content information to GAE, like the effect of the residual network.
Multiple layers are usually stacked in the deep network, and each layer captures different latent features of the data.
To combine the high-order relationship of the data (in GAE) with the potential details of the corresponding content information (in auto-encoder) layer-by-layer, we propose a cross-attention fusion mechanism, which highlights the discriminative information for clustering tasks.
Different from the traditional attention mechanism, our cross-attention fusion mechanism fuses two kinds of heterogeneous representations, i.e., the regular data and the irregular graph.



In this paper, we propose a novel clustering framework, named Cross-Attention Fusion based Enhanced Graph Convolutional Network (CaEGCN).
In CaEGCN, we can extract the high-order relationship between the data through the Graph Convolutional Auto-encoder module (GAE). 
To alleviate the over-smoothing problem of GAE and supplement the content information to GAE, we build a Content Auto-encoder module (CAE) composed of a common auto-encoder, which extracts the content information of the data.
Besides, we propose a cross-attention fusion mechanism to encode above two modules to output a complete representation.
In order to guide the optimal clustering direction of the entire model in an end-to-end manner, we introduce a self-supervised module.

The contributions of this paper are listed as follows as a summary,
\begin{itemize}
    \item We propose an end-to-end cross-attention fusion based deep clustering framework, in which the cross-attention fusion module creatively concatenates the graph convolutional auto-encoder module and content auto-encoder module in multiple layers;
    \item We propose a cross-attention fusion module to
    assign the attention weights to the fused heterogeneous representation; 
    \item In the graph convolutional auto-encoder module, we propose simultaneously reconstructing the content and relationship between the data, which effectively strengthens the clustering performance of CaEGCN;
    \item We test CaEGCN on the natural language, human behavior and image datasets to prove the robustness of CaEGCN.
\end{itemize}

The rest of the paper is organized as follows. In Section \ref{Sec:2}, we briefly review the graph convolutional network, deep clustering and attention mechanism, respectively. In Section \ref{Sec:3}, we detail the cross-attention fusion based enhanced graph convolutional network for clustering by presenting the four main modules. In Section \ref{Sec:4}, the proposed method is evaluated on clustering problems with several public datasets. Finally, the conclusion and the future work are discussed in Section \ref{Sec:5}.

\begin{figure*}[htbp]
    \centering
    \includegraphics[width=0.8\textwidth]{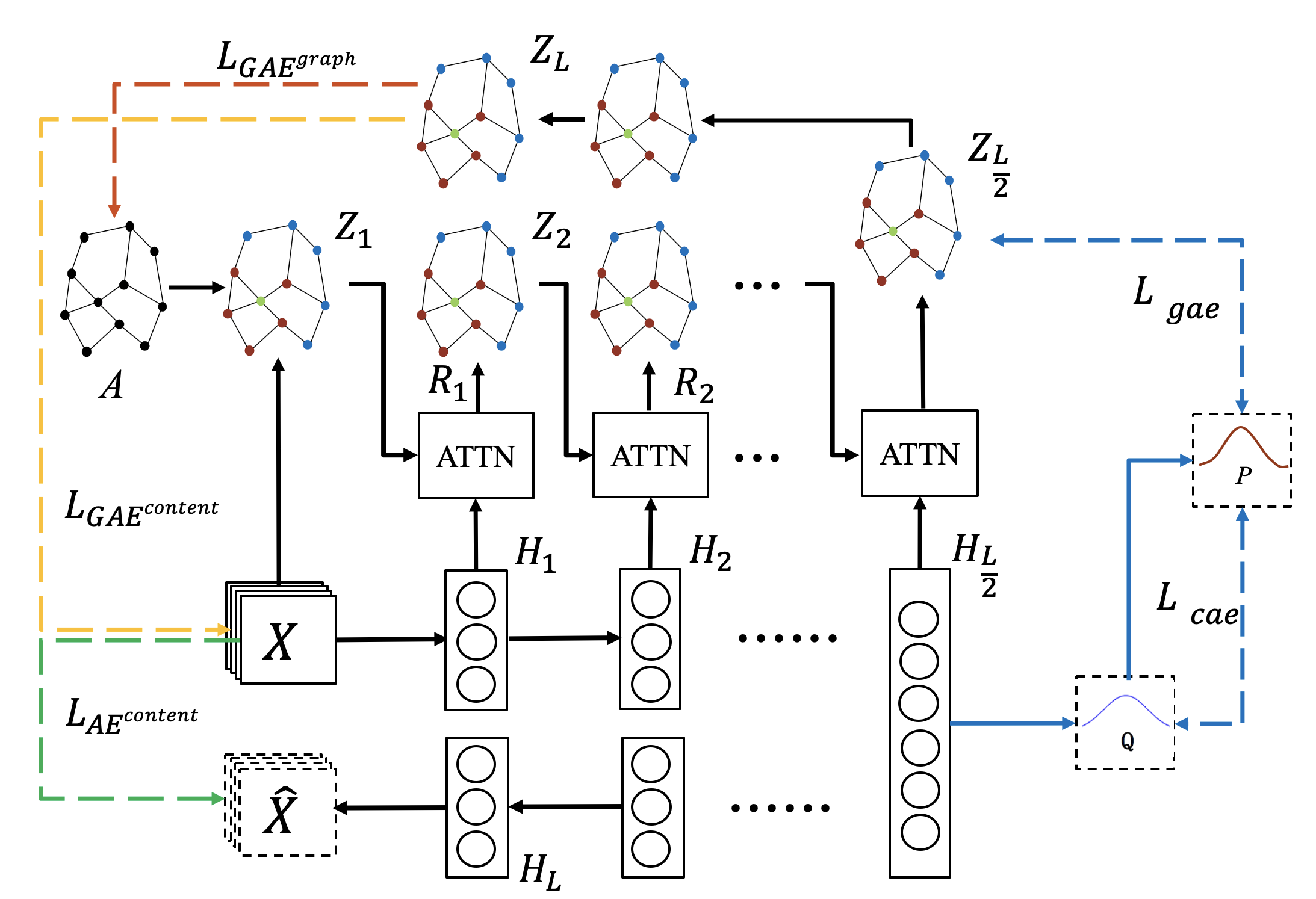}
    \caption{The conceptual framework of CaEGCN, which includes four modules: content auto-encoder module, graph convolutional auto-encoder module, cross-attention fusion module and self-supervised module.
    $X$ is the original data, $\hat{X}$ is the reconstructed data, and $A$ is the original graph. $H_l$ and $Z_l$ represent the $l$-th layer output of CAE and GAE, respectively.
    $R_l$ is the cross-attention fused representation of $H_l$ and $Z_l$.
    $\mathcal L_{{CAE}^{content}}$ is the content reconstruction loss of CAE. $\mathcal L_{{GAE}^{content}}$ and $\mathcal L_{{GAE}^{graph}}$ are the content reconstruction loss and graph reconstruction loss of GAE, respectively.
    $\mathcal L_{cae}$ and $\mathcal L_{gae}$ form the self-supervised module losses.}
    \label{overall}
\end{figure*}

\section{Related Work}\label{Sec:2}

In this section, we review the necessary knowledge related to the research of this paper, which are \emph {graph convolutional network, deep clustering and attention mechanism.}

\subsection{Graph Convolutional Network (GCN)}
Many research fields consider certain natural graph structures, such as the traffic road network \cite{zhao2020t}, the human skeleton points \cite{yan2018spatial} and molecules structures in biology \cite{sanyal2020proteingcn}.
Graph is a kind of irregular structural data, which is dispersive and disorderly. To cope with such irregular data, a lot of GCN based methods have been proposed.
These methods can be divided into two main categories: spectral-based GCN methods \cite{defferrard2016convolutional,kipf2017semi,bruna2014spectral} and spatial-based GCN methods \cite{velickovic2018graph,hamilton2017inductive}.
In a GCN, nodes can be assigned to the features of data, and the edge weight information describes the similarity between nodes, which shows that graph has a strong information organization ability.

The spectral-based GCN methods exploit the spectrum representation of a graph.
Kipf \emph{et al.} \cite{kipf2017semi} initially proposed the graph convolutional networks for prediction tasks, which simulates the graph convolutional operation through the local first-order approximation of spectral convolutions.
Wang \emph{et al.} \cite{wang2018graphgan} introduced the generative adversarial mechanism into the learning of graph representation, and developed a new graph softmax function utilizing the latent structure information of the data.

The spatial-based GCN methods directly define the operations on the graph and extract the information from the spatial neighbor groups.
Velickovic \emph{et al.} \cite{velickovic2018graph} proposed a graph attention network, which computes the corresponding hidden information for each node and
uses the attention mechanism to weight the importance of each node compared with its neighbors.
More comprehensive reviews about GCN can be found in \cite{wu2020a}.


\subsection{Deep Clustering}
The current deep learning researches mainly concentrate on the supervised learning tasks.
How to extend it onto a framework for unsupervised clustering is a meaningful problem.
Fortunately, some researchers have conducted the related works.
Xie \emph{et al.} \cite{xie2016unsupervised} proposed a deep embedded clustering method, which exploits the deep learning to learn the feature representations and the  cluster assignments of the data.
Ji \emph{et al.} \cite{ji2017deep} constructed a self-expression layer between the encoder and decoder of the auto-encoder. 

With the development of multi-view clustering, more and more researchers introduce the relationship between the data to enhance the clustering performance.
Kipf \emph{et al.} \cite{kipf2016variational} used the graph convolutional encoder and an inner product decoder to build a Variational Graph Auto-encoder (VGAE), which learns the latent features of undirected graphs for clustering.
Pan \emph{et al.} \cite{pan2020learning} improved the VGAE framework and introduced an adversarial regularization rule to optimize the learned representation for clustering.
Wang \emph{et al.} \cite{wang2019attributed} employed the graph attention network to weight the importance of neighboring nodes, and obtained a more accurate representation of each node.
Li \emph{et al.} \cite{li2020embedding} jointed the advantages of K-means and spectral clustering, and embedded it into the graph auto-encoder to generate the better data representations. 
Bo \emph{et al.} \cite{bo2020structural} transferred the representation learned by the auto-encoder to the corresponding GCN, and proposed a dual self-supervision mechanism to unify these two different deep neural architectures, which is an important baseline in this paper.

In above, GCN-based clustering methods update network parameters by reconstructing the adjacency matrix and sufficiently exploit the structure information, but they ignore the node information and the over-smoothing problem.

\subsection{Attention Mechanism}
Recently, in the fields of machine translation \cite{vaswani2017attention,tang2018why}, semantic segmentation \cite{fu2019dual} and image generation \cite{xu2015show}, the attention mechanism has become a trick module that improves the effectiveness of most models.
Self-attention is a variant of the traditional attention mechanism.
Vaswani \emph{et al.} \cite{vaswani2017attention} proposed the self-attention mechanism in machine translation applications, which obtains the satisfactory experimental results.
Besides, self-attention mechanism is robust and easily embedded into recurrent neural networks \cite{cheng2016long}, generative adversarial networks \cite{zhang2019self} and other neural networks, which also achieves the excellent experimental results.

Many scholars continuously optimize the self-attention mechanism.
Wang \emph{et al.} \cite{wang2019self} introduced a dependency tree into the self-attention mechanism to represent the relationship between words. Yu \emph{et al.}
\cite{yu2018qanet} extracted the local information through the convolution model to complement the global interaction of the self-attention mechanism. Xue \emph{et al.} \cite{xue2019danet} used the self-attention mechanism in image segmentation, which better achieves the accurate segmentation through the long-range context relations


To simultaneously handle the heterogeneous data in the proposed model, i.e., the regular data and the irregular graph, we propose the cross-attention fusion module in this paper. 

\section{Cross-Attention Fusion based Enhanced Graph Convolutional Network}\label{Sec:3}

In this section, we present a novel cross-attention fusion based enhanced graph convolutional network model, which sufficiently integrates the content information and the relationship between the data in a multi-level adaptive manner to improve the clustering performance.

The overall network architecture is shown in Figure \ref{overall}, consisting of four main modules: an auto-encoder module for extracting the content information; a GCN based auto-encoder module for exploiting the relationship between the data; a cross-attention module for concatenating above two modules,
 where the multi-level adaptive fusing strategy supplements the effective content information as much as possible during the transmission process; and a self-supervised module used to constrain the consistency of the distributions of middle layer representations.

\subsection{Constructing the Graph}
Before presenting the proposed CaEGCN model, we first construct the necessary graph of raw data.
Given a set of data $X\in\mathbb{R}^{D \times N}$ containing $N$ samples and the $i$-th sample $x_i\in \mathbb R^D$,
we employ the commonly-used $K$-nearest neighbor (KNN) to construct the corresponding graph to exhibit its structure information.

For the image data, we calculate the similarity between samples using the heat kernel method as  \cite{grigoryan2012heat},
\begin{equation}\label{Heat Kernel}
\begin{aligned}
    S_{ij}=e^{-\frac{{\|x_i- x_j\|_2^2}}{t}},
\end{aligned}
\end{equation}
where $t$ represents the variance scale parameter. 

As for the natural language data, the inner-product method is chosen to measure the similarity between samples as follow,
\begin{equation}
    \label{Dot-product}
    S_{ij}=  x_j^T  x_i.
\end{equation}

Then, with the above calculated similarities among all samples, we pick up the $K$ highest correlation neighbors of each sample and connect them; so, a graph $A$ is obtained. In many applications, the graph information actually comes with the given dataset $X$.

\subsection{Content Auto-encoder Module (CAE)}
As we know, deep convolutional network can effectively extract the critical features from the complex data. Auto-encoder can reconstruct the samples and reduce the missing information during the learning procedure, which is naturally proper for unsupervised learning.
To extract the content information in the data, we first train a deep convolutional network based auto-encoder module, which is named as Content Auto-encoder Module (CAE).

We represent the input of the $l$-th layer as $H_{l-1}$, then its output $H_l$ can be obtained by,
\begin{equation}
    \label{hidden_layer}
    \begin{aligned}
    &H_l=a_l(U_lH_{l-1}+b_l),\\
    &l=1,2,\cdots, L,
    \end{aligned}
\end{equation}
where the activation function of the $l$-th layer, $a_l$, can be chosen according to the practical applications, such as ReLU or Sigmoid.
$U_l$ and $b_l$ denote the weight and bias of the $l$-th layer of CAE, respectively.
In addition, the input in the first layer of CAE is the raw data $X$, i.e., $H_0= X$.
The  output in the final layer reconstructs the raw data, i.e., $\hat{X}=H_{L}$, and the final loss function of CAE can be defined as,
\begin{equation}
    \label{L_content}
    \begin{aligned}
    \mathcal L_{CAE^{content}}=\frac{1}{2}||X-\hat{X}||_F^2,
    \end{aligned}
\end{equation}
where $\|\cdot\|_F^2$ denotes the Frobenius norm.

\subsection{Cross-Attention Fusion Module}
As shown in Figure \ref{overall}, CAE extracts the content information in the data, and GAE exploits the corresponding relationship between the data. How to fuse these two kinds of information for clustering tasks is a key problem.

Cross-attention fusion mechanism has the global learning ability and good parallelism, which can further highlight the critical information in the fusion representations while suppressing the useless noise. 
Therefore, we use the cross-attention fusion mechanism to integrate the content information learned by CAE and the data relationship learned by GAE in a multi-level adaptive manner, which is the so-called Cross-Attention Fusion Module.

We define the cross-attention fusion mechanism as,
\begin{equation}\label{CrossAttention}
    \begin{aligned}
     R = F_{att}(Q,K,V),
    \end{aligned}
\end{equation}
where the query $Q=W^qY$, the key $K=W^kY$ and the value $V=W^vY$. The raw fusion representation $Y$ is the input of the cross-attention fusion module, which is defined as,
\begin{equation}\label{Y}
    \begin{aligned}
     Y=\gamma Z_l+(1-\gamma)H_l
    \end{aligned}
\end{equation}
where $H_l$ is the output of $l$-th layer in CAE and $Z_l$ is the output of corresponding layer in GAE. $\gamma$ is a trade-off parameter, which is set to $0.5$ in our experiments.


To discover the latent relationship between data and generalize the cross-attention fusion mechanism \eqref{CrossAttention}, we firstly calculate the similarity $s_{ab}$ between the fusion query $q_a$ and the fusion key $k_b$,
\begin{equation}
    \label{s}
    \begin{aligned}
       s_{ab}=q_a*k_b,
    \end{aligned}
\end{equation}
where $q_a$ and $k_b$ denote the $a$-th and $b$-th vectors in $Q$ and $K$, respectively.

Then, we execute the softmax normalization on above $s_{ab}$ to obtain the relevance weight $\alpha_{ab}$ as follows,
\begin{equation}
    \label{alpha}
    \begin{aligned}
    \alpha_{ab}= \text{softmax}(s_{ab}) =\frac{\exp(s_{ab})}{\sum^{D_{att}}_{a=0}\exp(s_{ab})}.
    \end{aligned}
\end{equation}




Finally, the output of the cross-attention fusion mechanism $R=(r_1, r_2, r_3, \cdots, r_N)$, i.e., the fusion representation of the data content information and the relationship between data, can be written as,
\begin{equation}
    \label{C}
    \begin{aligned}
    r_a=\sum_{b=0}^N\alpha_{ab}v_b.
    \end{aligned}
\end{equation}


To further perceive different aspects of the data, multi-head mechanism is also introduced, which contains multiple parallel cross-attention fusion modules. Specifically, we repeatedly project the query $Q$, key $K$ and value $V$ to obtain $M$ parallel cross-attention modules. 
Each cross-attention fusion module is regarded as one head, and each head has the different weight matrices $\{W^q_m \in \mathbb{R}^{N\times D_l}$, $W^k_m\in \mathbb{R}^{N\times D_l}$, $W^v_u\in \mathbb{R}^{N\times D_l}\}$ to linearly transform the fusion features $Q_m=W^q_mQ$, $K_m=W^k_mK$, $V_m=W^v_mV$ where $D_l$ is the dimensionality of the $l$-th layer. The $m$-th head is,
\begin{equation}
    \label{head}
    \begin{aligned}
    R^m = F_{att}(Q_m,K_m,V_m), m=1,2,3,\cdots, M.
    \end{aligned}
\end{equation}

We concatenate the outputs of all $M$ heads, and multiply the weight matrix $W\in  \mathbb{R}^{N \times(M\times D_l)} $ to get the final cross-attention fusion representation,
\begin{equation}
    \label{multihead}
    \begin{aligned}
    R = W \cdot Concat(R^1, \cdots, R^M),
    \end{aligned}
\end{equation}
where $Concat(\cdot)$ denotes the matrix concatenate operation. This is the so-called multi-head mechanism and cross-attention fusion module.

\subsection{Graph Convolutional Auto-Encoder Module (GAE)}
As we mentioned before, the relationship between the data can effectively improve the clustering performance. Most deep clustering methods only consider the content information of data, while ignoring the important data relationship \cite{xie2016unsupervised}.
Fortunately, Graph Convolutional Network (GCN) \cite{hamilton2017inductive} is able to handle such relationships and the content information of data collaboratively. 
To exploit GCN in unsupervised clustering tasks, we propose a GCN based Auto-Encoder module (GAE), which creatively reconstructs both graph and content information.

The previous cross-attention mechanism module combines the content representation $H_l$ from CAE with the relationship representation $Z_l$ 
from GAE to output a fusion representation $R_l$ in different layers. Then, the GAE executes the spectral graph convolution on such $R_l$ to learn the high-order discriminative information based on the adjacency matrix $A$. Finally, the middle layer $Z_{\frac{L}{2}}$ is used for clustering.


The convolution operation in each GAE layer can be expressed as follow,
\begin{equation}
    \label{GCN}
    Z_l=GAE(R_{l-1},A)=a_l(\hat{D}^{-\frac{1}{2}}\hat{A}\hat{D}^{-\frac{1}{2}}R_{l-1}U_{l}),
\end{equation}
where $\hat{D}^{-\frac{1}{2}}\hat{A}\hat{D}^{-\frac{1}{2}}$ is the approximated graph convolutional filter and $\hat{D}$ is the degree matrix of $\hat{A}$, where $\hat{D}_{ii}=\sum_{j}\hat{A}_{ij}$. With the identity matrix $I$ and the adjacency matrix $A$, we use $\hat{A} = A+I$ to ensure the self-loop in each node. Additionally, $U_{l}$ denotes the weight of the $l$-th layer, and
$Z_l$ is the output of the $l$-th GAE layer.

It should be noted that the input of the first layer in GAE is slightly different. The first layer just uses the raw data $X$ instead of $R_0$ as input,
\begin{equation}
    \label{fristGCN}
    Z_1=GAE(X,A)=a_l(\hat{D}^{-\frac{1}{2}}\hat{A}\hat{D}^{-\frac{1}{2}}X U_{1}).
\end{equation}

After this multi-layer learning, the GAE encoder encodes both the raw relationship $A$ and the content $X$ into a useful representation $Z_{\frac{L}{2}}$.
In order to preserve more information, we set the graph reconstruction and content reconstruction errors as the loss functions of GAE.

\vspace{1ex}
\textbf{i) Graph Reconstruction Loss.} We choose a simple inner product operation to reconstruct the relationship between samples as \cite{kipf2016variational},
\begin{equation}
    \label{inner_product}
    \tilde{A}=Sigmoid(Z_L^TZ_L),
\end{equation}
where $Z_L$ is the output of the last GAE layer, and $\tilde{A}$ is the reconstructed adjacency matrix.
The loss of graph reconstruction can be defined as,
\begin{equation}
    \label{graphloss}
    \mathcal L_{GAE^{graph}}=||A-\tilde{A}||_F^2.
\end{equation}

By minimizing the error between $A$ and $\tilde{A}$, the GAE module may preserve more data relationship in the latent representation $Z_{\frac{L}{2}}$ to improve the clustering performance.

\vspace{1ex}
\textbf{ii) Content Reconstruction Loss.}
Except for the relationship between the data, we also constrain the GAE module to preserve enough content information, which has much difference with formula \eqref{L_content};
so, we creatively define its loss function as,
\begin{equation}
    \label{contentloss}
   \mathcal L_{GAE^{content}}=||X-Z_L||_F^2,
\end{equation}
where $Z_L$ is the output of the last layer in GAE, which has the same size with the raw data $X$.
In this way, the GAE encodes both the relationship and the content of samples into a discriminative representation for clustering. 


\subsection{Self-Supervised Module}
It is difficult to judge whether the learned representation $Z_{\frac{L}{2}}$ is optimally for clustering during the optimization procedure. We need to give an optimization target about clustering.

To solve this problem, we firstly get a set of initial cluster centers $\{\beta_c\}_{c=1}^C$ by performing K-means on $H_{\frac{L}{2}}$, where $C$ is the number of clusters. These cluster centers guide approximately the optimization direction for $Z_{\frac{L}{2}}$, which is the so-called Self-Supervised Module.

We use the Student’s $t$-distribution \cite{maaten2008visualizing} to calculate the similarity between the middle layer representation $H_{\frac{L}{2}}$ and the cluster centers $\beta_c$ as follow,
\begin{equation}
    \label{T}
    t_{ic}=\frac{(1+||h_i-\beta_c||^2)^{-1}}{\sum_{c=1}^{C}(1+||h_i-\beta_{c}||^2)^{-1}}
\end{equation}
where $h_i$ is the $i$-th sample representation of $H_{\frac{L}{2}}$.
And $t_{ic}$ measures the probability that the $i$-th sample is assigned to the $c$-th cluster, 
so $T=[t_{ic}]$ is the overall soft assignment distribution. 

Furthermore, the choice of target distribution directly determines the clustering quality. We believe that the high-confidence assignments in $T$ is reliable and can be used as the target distribution.
We raise $p_{ic}$ to highlight the role of high-confidence distribution, 
\begin{equation}
    \label{P}
    p_{i c}=\frac{{t_{i c}^2}/f_c}{\sum_{c=1}^C(t_{ i c}^2/f_{c})}
\end{equation}
where $f_c=\sum_i t_{i c}$ is the soft cluster frequency.
The distribution of $T$ and $P$ should be close to each other as follow,
\begin{equation}
    \label{KL_cae}
   \mathcal L_{cae} = \text{KL}(P||T) = \sum_i\sum_c p_{i c}\log\frac{p_{i c}}{t_{i c}}.
\end{equation}

Similarly, it is easy to construct a soft assignment distribution $Z$ for the representation $Z_{\frac{L}{2}}$, then we can use the target distribution $P$ to supervise the distribution $Z$ as,
\begin{equation}
    \label{KL_gae}
   \mathcal L_{gae} = \text{KL}(P||Z) = \sum_i\sum_c p_{i c}\log\frac{p_{i c}}{z_{i c}}.
\end{equation}
Now, the optimization goals of GAE and CAE are unified into a distribution $P$, which makes the learned representation more suitable for clustering tasks.

\subsection{Overall Loss Function}
The overall objective loss function of Cross-Attention Fusion based Enhanced Graph Convolutional Network (CaEGCN) can be summarized as,
\begin{equation}
\begin{aligned}
    \mathcal{L}_{overall} &=\mathcal L_{GAE^{graph}}+\mathcal L_{GAE^{content}}\\
    &+ \mathcal L_{CAE^{content}}+\mathcal L_{cae}+\mathcal L_{gae}.
\end{aligned}
\end{equation}

There are five items in the above objective function, including three reconstruction losses and two self-supervision losses,
which optimizes the data representations for clustering tasks from different perspectives.

After optimizing the above objective function, the local optimal representation $G_{\frac{L}{2}}$ is obtained. Then, we perform the softmax operation on $G_{\frac{L}{2}}$ to get the final clustering results, i.e.,
$\text{max}(\text{softmax}(G_{\frac{L}{2}}))$.

\section{Experimental}\label{Sec:4}
In this section, CaEGAN is evaluated on various type of public datasets, including natural language, human behavior and image datasets. We present the experimental settings and analysis below.

\begin{table*}
\centering
\resizebox{!}{55mm}{
\begin{tabular}{c|c|cccccccc}
\hline
Dataset  & Metric                                                       & \multicolumn{1}{l}{K-means}                                               & AE                                                                                            & IEDC                                                                                          & VGAE                                                                      & ARGA                                                                      & DAEGC                                                                     & SDCN                                                                                          & CaEGCN                                                                             \\ \hline
    ACM      & \begin{tabular}[c]{@{}c@{}}ACC\\ NMI\\ ARI\\ F1\end{tabular} & \begin{tabular}[c]{@{}c@{}}0.6820\\ 0.3263\\ 0.3119\\ 0.6846\end{tabular} & \begin{tabular}[c]{@{}c@{}}0.8278\\ 0.5020\\ 0.5553\\ 0.8295\end{tabular}                     & \begin{tabular}[c]{@{}c@{}}0.8645\\ 0.5824\\ 0.6421\\ 0.8632\end{tabular}                     & \begin{tabular}[c]{@{}l@{}}0.8294\\ 0.5285\\ 0.5618\\ 0.8286\end{tabular} & \begin{tabular}[c]{@{}l@{}}0.8327\\ 0.5039\\ 0.5646\\ 0.8335\end{tabular} & \begin{tabular}[c]{@{}l@{}}0.8694\\ 0.5618\\ 0.5935\\ 0.8707\end{tabular} & \begin{tabular}[c]{@{}c@{}}\underline{0.8860}\\ \underline{0.6326}\\ \underline{0.6931}\\ \underline{0.8857}\end{tabular}                     & \textbf{\begin{tabular}[c]{@{}c@{}}0.9012\\ 0.6703\\ 0.7300\\ 0.9009\end{tabular}} \\ \hline
DBLP     & \begin{tabular}[c]{@{}c@{}}ACC\\ NMI\\ ARI\\ F1\end{tabular} & \begin{tabular}[c]{@{}c@{}}0.3646\\ 0.0886\\ 0.0657\\ 0.2637\end{tabular} & \begin{tabular}[c]{@{}c@{}}0.5435\\ 0.2220\\ 0.1651\\ 0.5325\end{tabular}                     & \begin{tabular}[c]{@{}c@{}}0.6571\\ 0.3080\\ 0.3210\\ 0.6439\end{tabular}                     & \begin{tabular}[c]{@{}l@{}}0.5763\\ 0.2189\\ 0.2348\\ 0.5456\end{tabular} & \begin{tabular}[c]{@{}l@{}}0.5450\\ 0.2019\\ 0.1949\\ 0.5343\end{tabular} & \begin{tabular}[c]{@{}l@{}}0.6205\\ 0.3249\\ 0.2103\\ 0.6175\end{tabular} &
\begin{tabular}[c]{@{}c@{}} \underline{0.6613}\\ \underline{0.3249}\\ \underline{0.3338}\\ \underline{0.6556}\end{tabular}                     & \textbf{\begin{tabular}[c]{@{}c@{}}0.6823\\ 0.3388\\ 0.3617\\ 0.6669\end{tabular}} \\ \hline
Citeseer & \begin{tabular}[c]{@{}c@{}}ACC\\ NMI\\ ARI\\ F1\end{tabular} & \begin{tabular}[c]{@{}c@{}}0.3384\\ 0.1502\\ 0.0893\\ 0.2246\end{tabular} & \begin{tabular}[c]{@{}c@{}}0.5909\\ 0.3066\\ 0.3134\\ 0.5483\end{tabular}                     & \begin{tabular}[c]{@{}c@{}}0.6023\\ 0.3074\\ 0.2924\\ 0.5230\end{tabular}                     & \begin{tabular}[c]{@{}l@{}}0.5161\\ 0.2572\\ 0.2405\\ 0.4184\end{tabular} & \begin{tabular}[c]{@{}l@{}}0.5912\\ 0.3069\\ 0.3138\\ 0.5485\end{tabular} & \begin{tabular}[c]{@{}l@{}}\underline{0.6454}\\ \underline{0.3641}\\ \underline{0.3778}\\ \textbf{0.6220}\end{tabular} &
\begin{tabular}[c]{@{}c@{}} 0.6222\\ 0.3601\\0.3623\\ 0.5893\end{tabular}                     & \begin{tabular}[c]{@{}c@{}}\textbf{0.6802}\\ \textbf{0.4000}\\ \textbf{0.4240}\\ \underline{0.6138}\end{tabular} \\ \hline
HHAR     & \begin{tabular}[c]{@{}c@{}}ACC\\ NMI\\ ARI\\ F1\end{tabular} & \begin{tabular}[c]{@{}c@{}}0.5998\\ 0.5887\\ 0.4609\\ 0.5833\end{tabular} & \multicolumn{1}{l}{\begin{tabular}[c]{@{}l@{}}0.4621\\ 0.3610\\ 0.2257\\ 0.4182\end{tabular}} & \multicolumn{1}{l}{\begin{tabular}[c]{@{}l@{}}0.7920\\ 0.7960\\ 0.7033\\ 0.7333\end{tabular}} & \begin{tabular}[c]{@{}l@{}}0.6252\\ 0.6059\\ 0.4601\\ 0.5696\end{tabular} & \begin{tabular}[c]{@{}l@{}}0.7040\\ 0.7154\\ 0.6114\\ 0.6667\end{tabular} & \begin{tabular}[c]{@{}l@{}}0.7651\\ 0.6910\\ 0.6038\\ 0.7689\end{tabular} & \multicolumn{1}{l}{\begin{tabular}[c]{@{}l@{}} \underline{0.8449}\\ \underline{0.8021}\\ \underline{0.7292}\\ \underline{0.8297}\end{tabular}} & \textbf{\begin{tabular}[c]{@{}c@{}}0.8742\\ 0.8256\\ 0.7627\\ 0.8724\end{tabular}} \\ \hline
USPS     & \begin{tabular}[c]{@{}c@{}}ACC\\ NMI\\ ARI\\ F1\end{tabular} & \begin{tabular}[c]{@{}c@{}}0.6682\\ 0.6272\\ 0.5464\\ 0.6494\end{tabular} & \multicolumn{1}{l}{\begin{tabular}[c]{@{}l@{}}0.4402\\ 0.4850\\ 0.3082\\ 0.3665\end{tabular}} & \multicolumn{1}{l}{\begin{tabular}[c]{@{}l@{}}0.7684\\ 0.7795\\ 0.7011\\ 0.7565\end{tabular}} & \begin{tabular}[c]{@{}l@{}}0.6381\\ 0.7004\\ 0.5636\\ 0.5861\end{tabular}  & \begin{tabular}[c]{@{}l@{}}0.7196\\ 0.6859\\ 0.6081\\ 0.7093\end{tabular} & \begin{tabular}[c]{@{}l@{}}0.7355\\ 0.7112\\ 0.6333\\ 0.7245\end{tabular} & \multicolumn{1}{l}{\begin{tabular}[c]{@{}l@{}}\underline{0.7722}\\ \underline{0.7907}\\ \textbf{0.7110}\\ \underline{0.7626}\end{tabular}} & \begin{tabular}[c]{@{}c@{}}\textbf{0.7755}\\ \textbf{0.7923}\\ \underline{0.7107}\\\textbf{ 0.7634}\end{tabular} \\ \hline
\end{tabular}}\\[2mm]
\caption{Clustering results on all five datasets. We mark the best-performing and the second-best-performing results by bolded and underlined.}\label{results}
\end{table*}

\subsection{Datesets}
\begin{itemize}
    \item ACM \cite{bo2020structural} contains $3025$ papers of $3$ major categories (i.e., database, wireless communication and data mining). The keywords of each paper are chosen as its feature. Different papers of the same author should have the relative strong correlation, so we can construct the structure graph for GCN.
    \item DBLP \cite{ley2009dblp} is an author network dataset collected from the DBLP website, which includes $4057$ authors of $4$ categories. The research fields of each author are treated as the feature.
    \item Citeseer\footnote{https://csxstatic.ist.psu.edu/downloads/data} is a citation network dataset which is composed of paper features and citation connections between papers. This dataset has $3327$ papers of $6$ categories.
    \item HHAR \cite{stisen2015smart} 
    consists of $10299$ sensor records collected from smart phones and smart watches, which is divided into $6$ categories, including biking, sitting, standing, walking, stair up and stair down.
    \item USPS \cite{lecun2015deep} contains $9298$ gray images of $10$ hand-written digits, and the size of each images is $16 \times 16$.
\end{itemize}

\subsection{Compared Methods}
To verify the effectiveness of CaEGCN, we compare it with several state-of-the-art clustering methods, including,
\begin{itemize}
    \item \textbf{K-means} \cite{lecun2015deep} is a basic clustering algorithm based on the content of data only.
    \item \textbf{Auto-Encoder (AE)} \cite{hinton2006reducing} 
    performs K-means on the low-dimensional representations learned from the deep auto-encoder network.
    \item \textbf{Improved deep embedded  clustering (IEDC)} \cite{guo2017improved} adds the clustering oriented loss and the reconstruction loss to the deep auto-encoder network, which realizes the one-step clustering of low-dimensional representations.
    \item \textbf{Variational Graph Auto-Encoders (VGAE)} \cite{kipf2016variational} is a variational graph auto-encoder with both topology and content information, which introduces the GCN architecture and the graph reconstruction loss to build a graph convolutional auto-encoder network.
    \item \textbf{Adversarially Regularized Graph Auto-encoder (ARGA)} \cite{pan2020learning} is a GAN architecture deep clustering model. They firstly construct a graph convolutional auto-encoder network, then the adversarial training principle is applied to enforce the latent codes to match a prior Gaussian or uniform distribution.
    \item \textbf{Deep Attentional Embedded Graph Clustering (DAEGC)} \cite{wang2019attributed} uses the graph attention network to build the encoder and trains the internal product decoder to reconstruct the graph structure. In addition, soft labels are generated according to the graph embedding to monitor the self-training graph clustering process.
    \item \textbf{Structural Deep Clustering Network (SDCN)} \cite{bo2020structural} uses the structure information learned by the GCN module to strengthen the data representation learned by the auto-encoder. Furthermore, it constructs a dual self-supervised loss to combine two networks and supervise clustering, which is an important baseline.
\end{itemize}

The parameter settings of compared methods are listed below.
For K-means, we repeatly run K-means $20$ times and report the best result.
For AE and IEDC, following the work in \cite{bo2020structural}, the network dimension of each dataset is set as $500-500-2000-10-2000-500-500$.
VGAE is a two-layer network, so we set network dimension of its encoder as $500-10$, \cite{kipf2016variational}.
For ARGA, following the work in \cite{pan2020learning}, we set the dimension of the encoder as $32-16$, in addition, the discriminator's dimension is set as $16-64$.
For DAEGC, the dimension of the graph attention encoder is set as $256-16$, \cite{wang2019attributed}.
For SDCN, we set the dimensions of the encoder and GCN module for $500-500-2000-10$, \cite{bo2020structural}.

To evaluate all methods from multiple aspects, we choose four popular clustering evaluation metrics, including Accuracy (ACC), Normalized Mutual Information (NMI), Average Rand Index (ARI) and macro F1-score (F1). For all metrics, a higher score indicates a better clustering quality.

\subsection{Parameter Settings}
When a dataset does not come with graph information, to construct the initial graph of data, we select the popular K-Nearest-Neighbor algorithm (KNN), and the value of $K$ is positively correlated with the number of samples and categories.
Following the strategy in SDCN \cite{bo2020structural}, we tune different $K$ in the range \{$3,5,10$\} to get the best performance. Generally, USPS and HHAR employ $K=10$ and $K=5$ to construct the corresponding graphs, respectively. As for ACM, DBLP, and Citeseer, we directly exploit the existing graphs in the datasets.

In the proposed CaEGCN, we set the dimensions of both CAE and GAE modules as $input-500-10-cluster-500-500-output$, where $input$ and $output$ denote the dimension of the raw data, and $cluster$ represents the number of cluster categories.
The purpose of the last layer in the decoder is to reconstruct the raw data, so the dimension of the last layer equals to the first layer, i.e., $output=input$.

In the cross-attention fusion module, we set the number of heads as $8$.
In the self-supervised module, the iteration number of K-means is set as $1000$ to initialize the cluster centers.
At last, we employ the Xavier method to initialize our model parameters \cite{glorot2010understanding}, and the initial learning rate is set as $0.001$.

\begin{figure*}
\centering
\subfigure[ ACM-raw data.]{ \label{ACMR}
\includegraphics[width=0.22\linewidth]{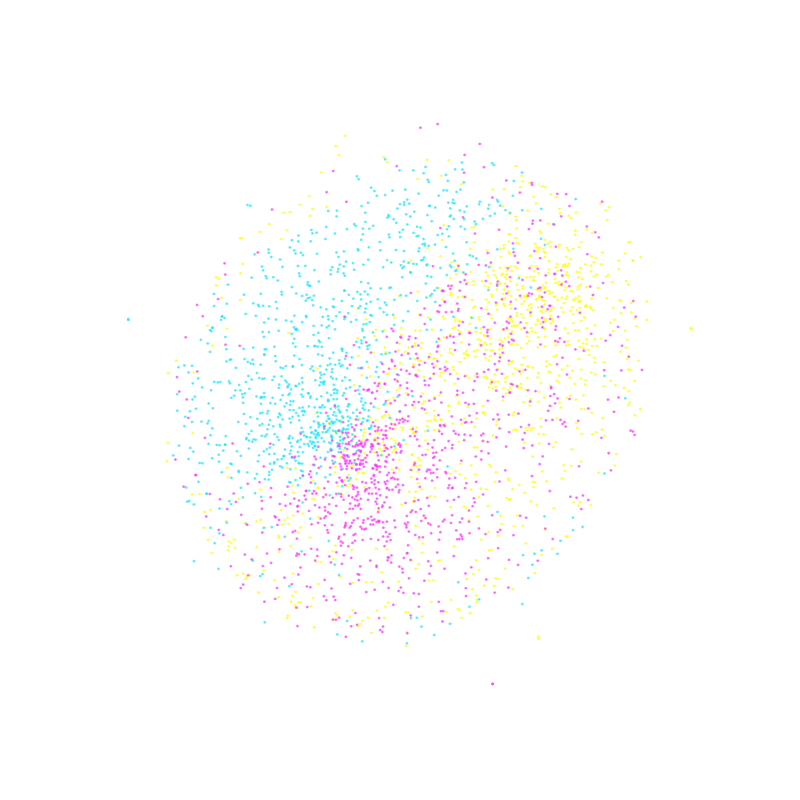}}
\subfigure[ACM-CaEGCN.]{ \label{ACMC}
\includegraphics[width=0.22\linewidth]{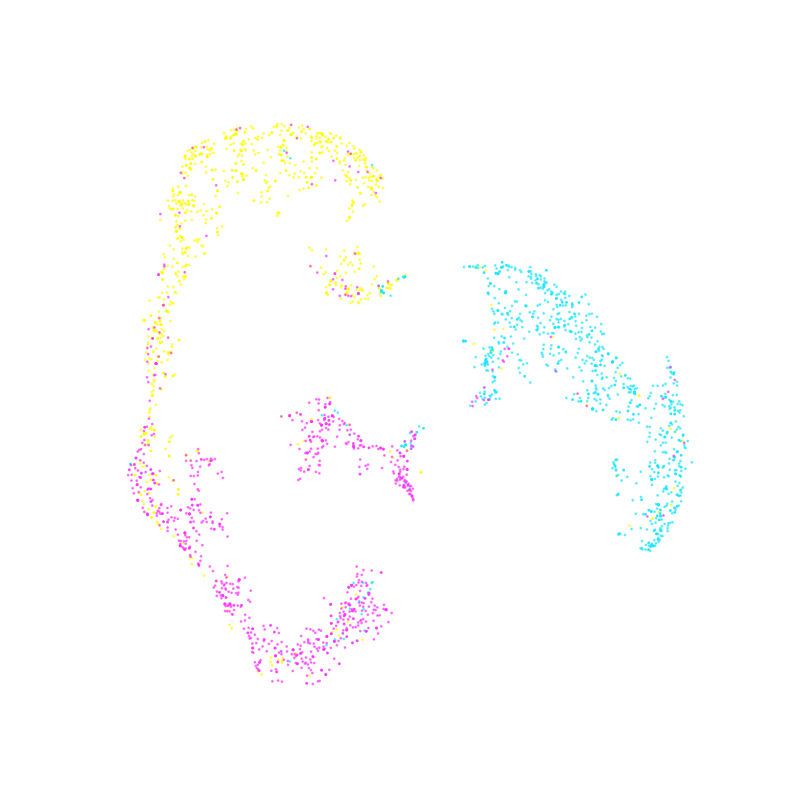}}
\subfigure[DBLP-raw data.]{ \label{DBLPR} 
\includegraphics[width=0.22\linewidth]{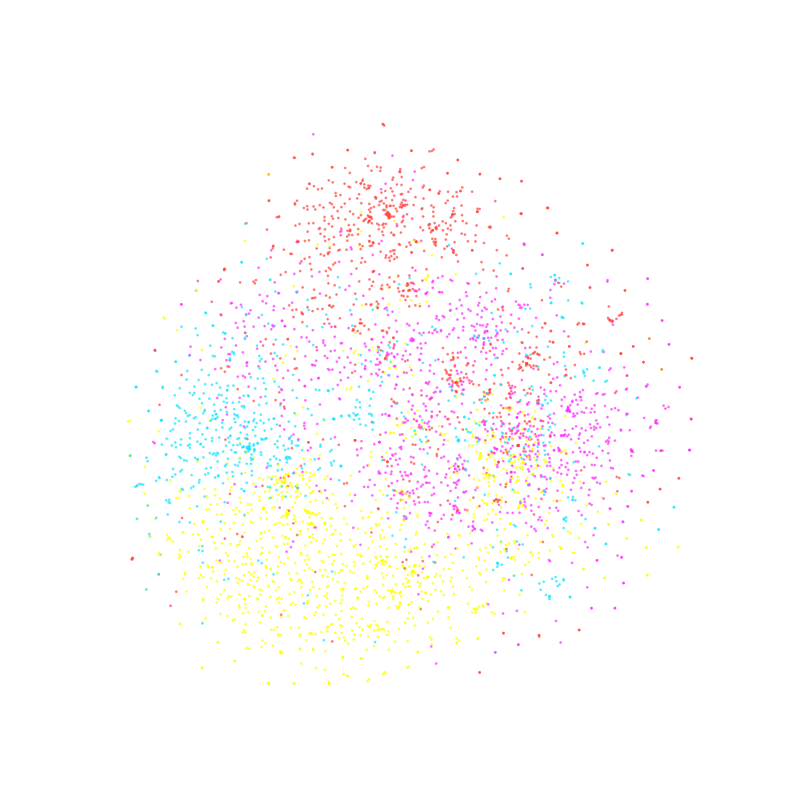}}
\subfigure[DBLP-CaEGCN.]{ \label{DBLPC} 
\includegraphics[width=0.22\linewidth]{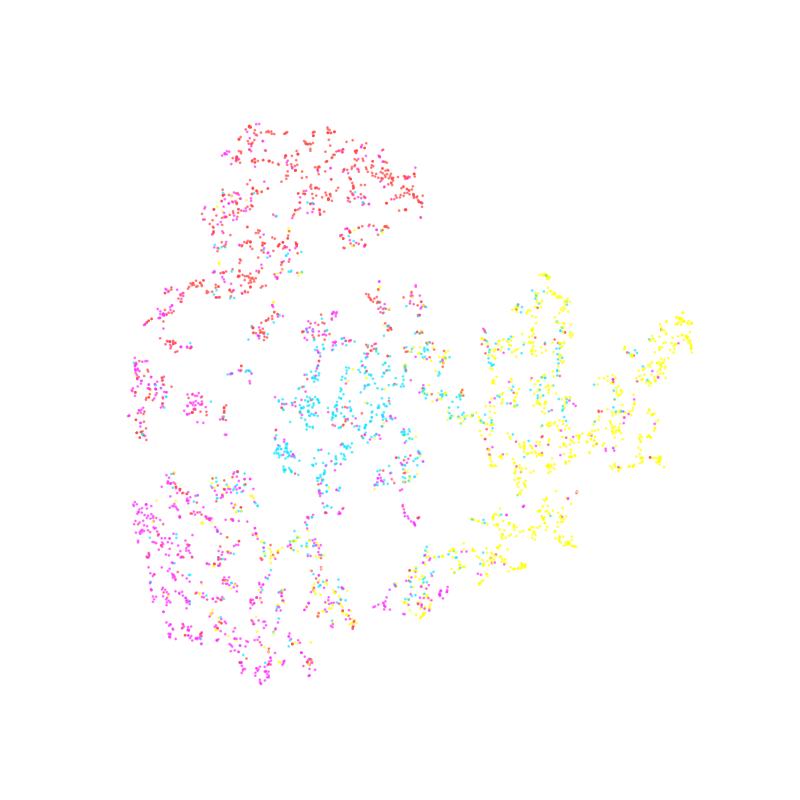}}
\subfigure[Citeseer-raw data.]{\label{CitesseerR}
\includegraphics[width=0.22\linewidth]{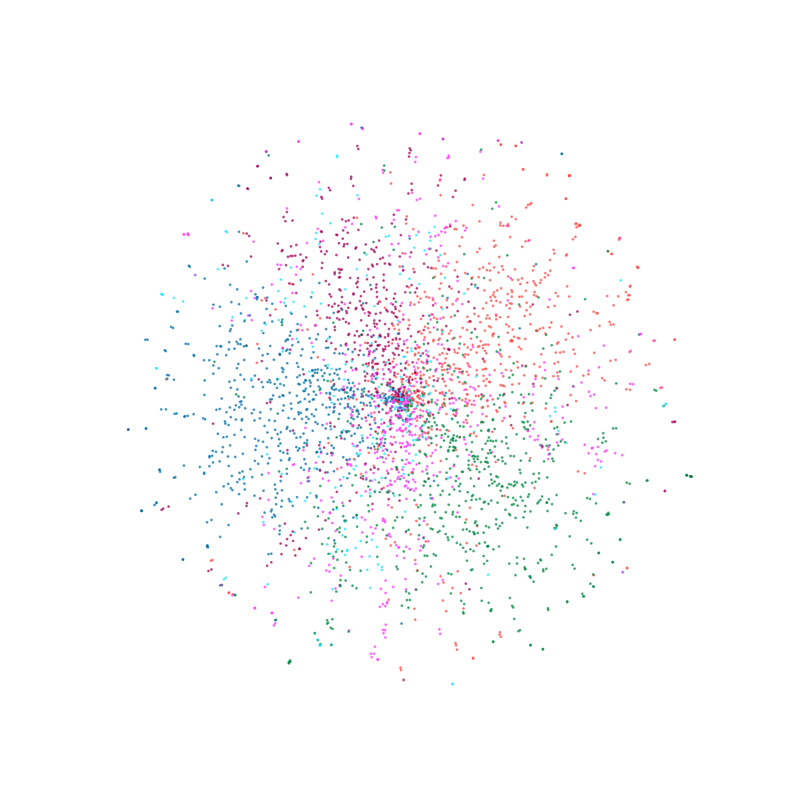}}
\subfigure[Citeseer-CaEGCN.]{\label{CitesseerC}
\includegraphics[width=0.22\linewidth]{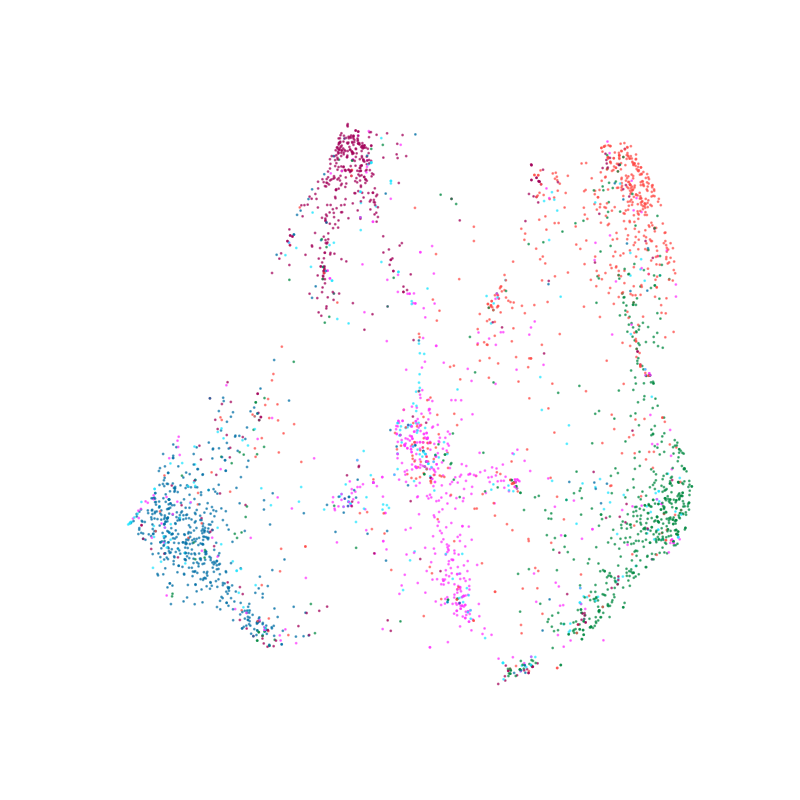}}
\subfigure[HHAR-raw data.]{ \label{HHARR} 
\includegraphics[width=0.22\linewidth]{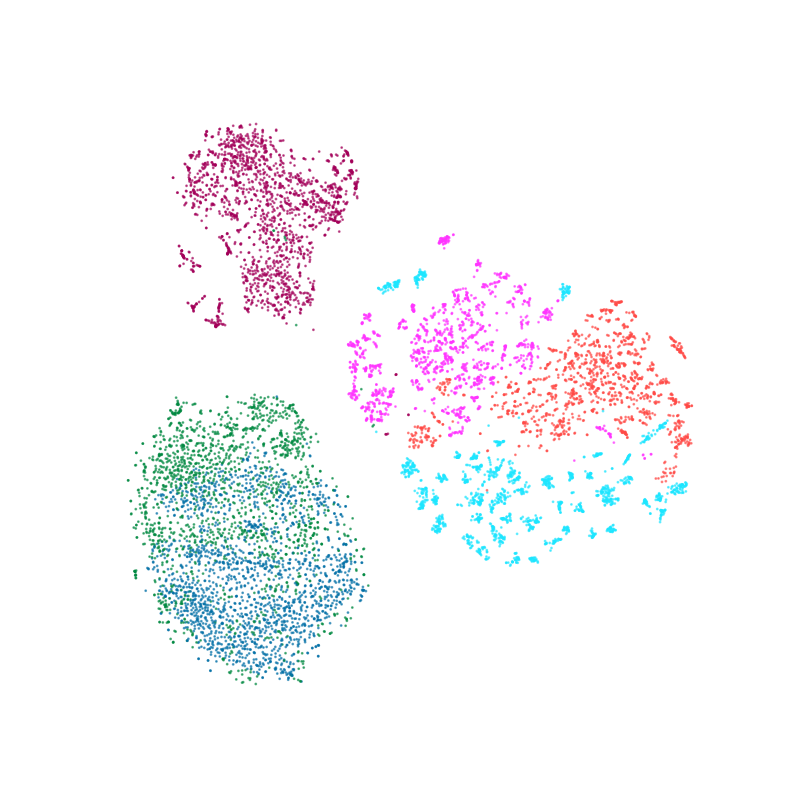}}
\subfigure[HHAR-CaEGCN.]{ \label{HHARC} 

\includegraphics[width=0.22\linewidth]{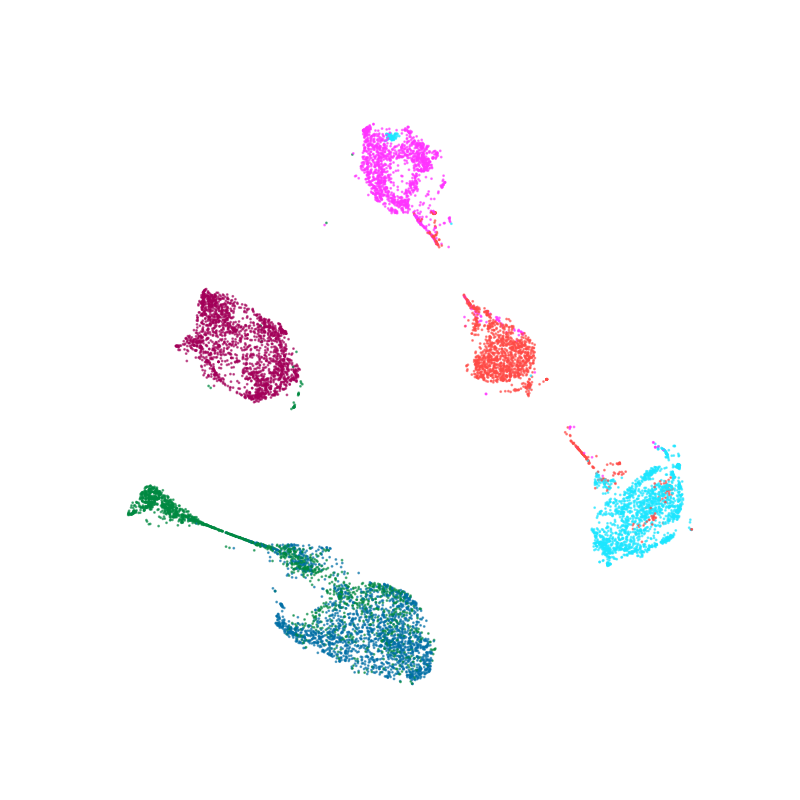}}
\subfigure[USPS-raw data.]{ \label{USPSR} 

\includegraphics[width=0.22\linewidth]{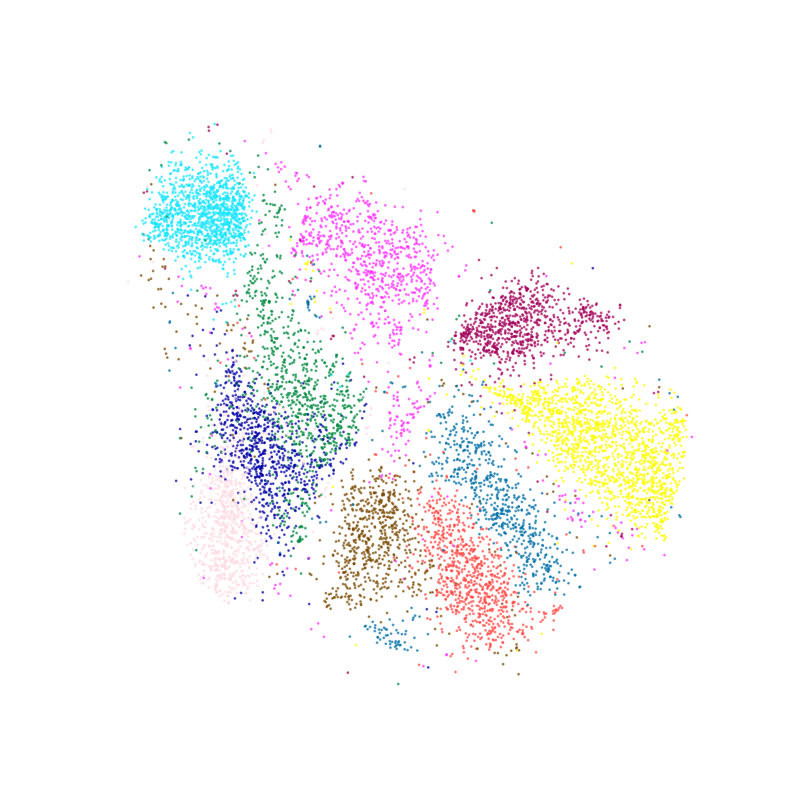}}
\subfigure[USPS-CaEGCN.]{ \label{USPSC} 

\includegraphics[width=0.22\linewidth]{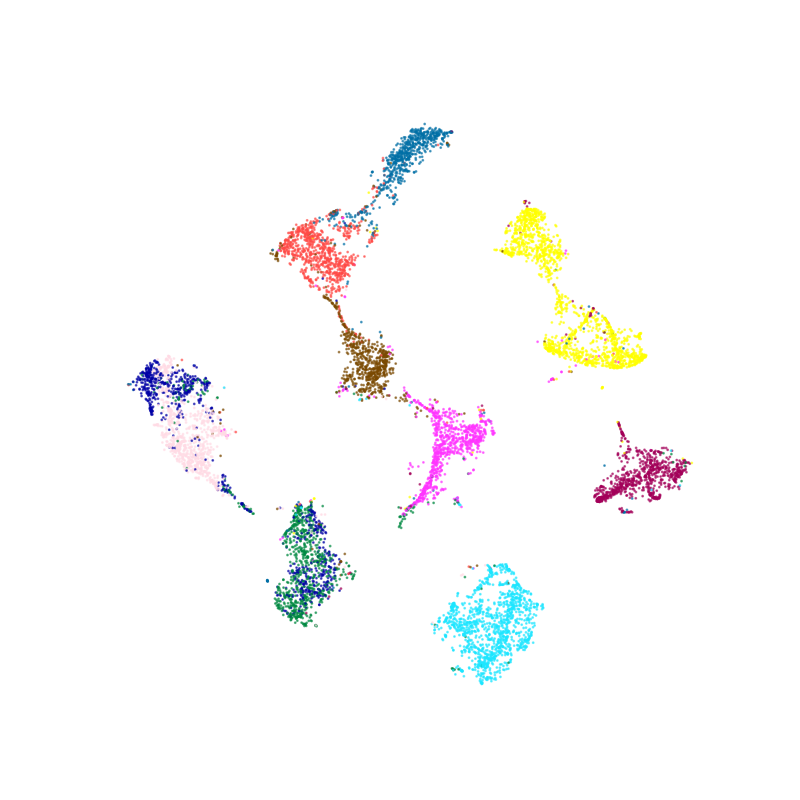}}
\caption{2D visualization. The comparison of the raw data and the clustering results of CaEGCN on ACM, DBLP, Citeseer, HAR and USPS datasets.} \label{Samplefig}
\end{figure*}

\subsection{Experiment Results Analysis}
Table \ref{results} exhibits the whole experimental results compared with other clustering methods. Obviously, the proposed CaEGCN model achieves the best performance in most cases.

We can see that the content information based deep clustering methods (such as AE and IEDC) work better than the graph convolutional auto-encoder method VGAE. The reason is that VGAE exists the over-smoothing problem. In other words, the information received by the nodes has a low signal-to-noise ratio.
SDCN and the proposed CaEGCN supplement the content information into the GCN module in each layer, which effectively relieves the over-smoothing problem, so SDCN and CaEGCN receive the satisfactory performance.

The experimental results show that SDCN and the CaEGCN are superior to other methods on the whole datasets. Compared with directly using GCN, supplementing the content information into the structure representation layer by layer can help the clustering work better, and it also illustrates the significance of the interaction between the heterogeneous information.
In addition, the proposed CaEGCN performs better than SDCN in most cases, which proves that the cross-attention fusion module in CaEGCN can promote the learned data representation containing more prominent information for clustering tasks.

\begin{table*}
\centering
\resizebox{!}{55mm}{
\begin{tabular}{r|c|cccc}
\hline
Dataset  & Metric                                                       & CaEGCN                                                                             & \begin{tabular}[c]{@{}c@{}}CaEGCN\\ w/o attention\end{tabular}            & \begin{tabular}[c]{@{}c@{}}CaEGCN\\ w/o graph\end{tabular}                & \begin{tabular}[c]{@{}c@{}}CaEGCN\\ w/o content\end{tabular}              \\ \hline
ACM      & \begin{tabular}[c]{@{}c@{}}ACC\\ NMI\\ ARI\\ F1\end{tabular} & \textbf{\begin{tabular}[c]{@{}c@{}}0.9012\\ 0.6703\\ 0.7300\\ 0.9009\end{tabular}} & \begin{tabular}[c]{@{}c@{}}0.8883\\ 0.6415\\ 0.6994\\ 0.8879\end{tabular} & \begin{tabular}[c]{@{}c@{}}0.8860\\ 0.6391\\ 0.6989\\ 0.8877\end{tabular} & \begin{tabular}[c]{@{}c@{}}0.8869\\ 0.6394\\ 0.6961\\ 0.8866\end{tabular} \\ \hline
DBLP     & \begin{tabular}[c]{@{}c@{}}ACC\\ NMI\\ ARI\\ F1\end{tabular} & \textbf{\begin{tabular}[c]{@{}c@{}}0.6823\\ 0.3388\\ 0.3617\\ 0.6669\end{tabular}} & \begin{tabular}[c]{@{}c@{}}0.6766\\ 0.3261\\ 0.3515\\ 0.6581\end{tabular} & \begin{tabular}[c]{@{}c@{}}0.6628\\ 0.3127\\ 0.3257\\ 0.6530\end{tabular} & \begin{tabular}[c]{@{}c@{}}0.6732\\ 0.3249\\ 0.3422\\ 0.6620\end{tabular} \\ \hline
Citeseer & \begin{tabular}[c]{@{}c@{}}ACC\\ NMI\\ ARI\\ F1\end{tabular} & \textbf{\begin{tabular}[c]{@{}c@{}}0.6802\\ 0.4000\\ 0.4240\\ 0.6138\end{tabular}} & \begin{tabular}[c]{@{}c@{}}0.6411\\ 0.3550\\ 0.3706\\ 0.5975\end{tabular} & \begin{tabular}[c]{@{}c@{}}0.6616\\ 0.3905\\ 0.3898\\ 0.5744\end{tabular} & \begin{tabular}[c]{@{}c@{}}0.6267\\ 0.3533\\ 0.3463\\ 0.5406\end{tabular} \\ \hline
HHAR     & \begin{tabular}[c]{@{}c@{}}ACC\\ NMI\\ ARI\\ F1\end{tabular} & \textbf{\begin{tabular}[c]{@{}c@{}}0.8742\\ 0.8256\\ 0.7627\\ 0.8724\end{tabular}} & \begin{tabular}[c]{@{}c@{}}0.8637\\ 0.8090\\ 0.7320\\ 0.8595\end{tabular} & \begin{tabular}[c]{@{}c@{}}0.8693\\ 0.8231\\ 0.7599\\ 0.8624\end{tabular} & \begin{tabular}[c]{@{}c@{}}0.8650\\ 0.8224\\ 0.7559\\ 0.8568\end{tabular} \\ \hline
USPS     & \begin{tabular}[c]{@{}c@{}}ACC\\ NMI\\ ARI\\ F1\end{tabular} & \textbf{\begin{tabular}[c]{@{}c@{}}0.7755\\ 0.7871\\ 0.7107\\ 0.7634\end{tabular}} & \begin{tabular}[c]{@{}c@{}}0.7623\\ 0.7712\\ 0.6902\\ 0.7552\end{tabular} & \begin{tabular}[c]{@{}c@{}}0.7698\\ 0.7746\\ 0.6984\\ 0.7566\end{tabular} & \begin{tabular}[c]{@{}c@{}}0.7704\\ 0.7794\\ 0.7019\\ 0.7578\end{tabular} \\ \hline
\end{tabular}}\\[2mm]
\caption{The results of ablation experiments on all five datasets. We mark the best-performing result by bolded.}\label{AblationResult}
\end{table*}

For the academic papers datasets, various factors interfere the clustering or recognition tasks, such as the cross-domain applications of popular algorithms, different research topics in the same field and different research fields of the same author, and so on.
As for the ACM dataset,
the CaEGCN achieves the significant improvements in all four evaluation metrics. The accuracy of CaEGCN increases by from $1.7\%$ compared with SDCN to $24.1\%$ compared with K-means; The NMI of CaEGCN increases from $5.6\%$ compared with SDCN to $51.9\%$ compared with K-means; The ARI of CaEGCN increases by from $5\%$ compared with SDCN to $57.3\%$ compared with K-means; The F1 score of CaEGCN increases by from $1.7\%$ compared with SDCN to $24\%$ compared with K-means.

We also note that VGAE has achieved good clustering results, which simultaneously considers the graph topology information and node content of the graph.
ARGA uses the adversarial training to optimize the method based on graph convolution, and achieves some improvements. DAEDC uses the graph attention module, and also achieves the better experimental results. The huge gap between the CaEGCN and SDCN (and others) further proves the superiority of the CaEGCN. Similar to the ACM dataset, DBLP and Citeseer are datasets related to academic papers, and their experimental results show the same pattern and trend.

Large scale dataset is an important challenge for clustering methods. When the scale of samples increases, the performance of many state-of-the-art clustering methods drops dramatically. The scale of HHAR and USPS datasets are $3$ times larger than the previous datasets.

For the HHAR dataset, it is difficult to distinguish some human daily behaviors, 
e.g., walking and biking. This imposes a challenge for clustering tasks. Compared with VGAE which integrates the structural features into the content information, the accuracy of the CaEGCN increases by $28.48\%$; compared with ARGA, the accuracy increases by $19.47\%$; and compared with DAEGC, the accuracy increases by $12.48\%$.
It is our belief that the poor performance of these methods is due to 
the over-smoothing problem of GCN.  The experimental results prove the effectiveness of our cross-attention module.
Compared with the best baseline SDCN, the accuracy of CaEGCN still increases by $3.4\%$.

For the USPS dataset, its background is simple, so we regard it as a baseline data to test the robustness of the CaEGCN.
The experimental results of the CaEGCN, SDCN, and IDEC are similar. This may be caused by the fact that 
the content information of the some handwritten digit images is difficult to distinct. 
We emphasize here that each node in the initial graph we constructed can only connect $10$ closest nodes when considering efficiency, so the initial graph fails to contain enough valuable relationship between the data. Such a limited graph restricts the learning ability of the convolutional network.

In summary, these performance improvements of the CaEGCN can be attributed to two aspects: first, the cross-attention fusion mechanism integrates the content information and the data relationship, and highlights the critical information in the fusion representations;
second, the self-supervised module further optimizes the distributions of the middle layer representations to strengthen the performance of deep clustering.

\subsection{Clustering Result Visualization}
We visualize the clustering results of five datasets in a two-dimensional space with the t-SNE algorithm \cite{maaten2008visualizing}. The location distribution of the raw data is overlapping, while the CaeGCN can obviously drive the raw data into different groups.

\subsection{Ablation Experiment Analysis}
To prove the effectiveness of each critical module in our model, we designed a set of ablation experiments. Specifically, we repeatedly remove one module from the CaEGCN model and test the clustering performance of these incomplete models on five datasets.
The designed incomplete models are displayed as follow,
\begin{itemize}
    \item \textbf{CaEGCN w/o attention:} The proposed CaEGCN without the cross-attention fusion module.
    \item \textbf{CaEGCN w/o graph:} The proposed CaEGCN without the graph reconstruction loss $\mathcal L_{GAE^{graph}}$ in the GAE module,
    \begin{equation}
      \begin{aligned}
    \mathcal{L}_{overall} &=\mathcal L_{GAE^{content}} + \mathcal L_{CAE^{content}}+\mathcal L_{cae}+\mathcal L_{gae}.
      \end{aligned}
    \end{equation}
    \item \textbf{CaEGCN w/o content:} The proposed CaEGCN without the content reconstruction loss $\mathcal L_{GAE^{content}}$ in the GAE module,
    \begin{equation}
      \begin{aligned}
    \mathcal{L}_{overall} &=\mathcal L_{GAE^{graph}} + \mathcal L_{CAE^{content}}+\mathcal L_{cae}+\mathcal L_{gae}.
      \end{aligned}
    \end{equation}
\end{itemize}

From Table \ref{AblationResult}, we observe that the CaEGCN still achieves the best results on all datasets. The clustering results of the above three incomplete models decline, which verifies the importance of each critical module.

Among them, the experimental results of \textbf{CaEGCN w/o attention} drop sharply, which proves that the lack of the content and data relationship fusion can decrease the learning ability of the GAE module.

Without the graph reconstruction loss $\mathcal L_{GAE^{graph}}$, the clustering performance of \textbf{CaEGCN w/o graph} still decrease obviously. This reconstruction loss ensures the learned middle layer representations have abundant structure information, which improves the clustering performance.
Meanwhile, the experimental results of the \textbf{CaEGCN w/o graph} on the datasets using the original graph (e.g., ACM, DBLP) significantly decrese, which reflects that the graph reconstruction loss can effectively improve the quality of data representations with the accurate graph structure.

As for \textbf{CaEGCN w/o content}, its experimental results are not bad.
Compared with other two modules, the impact of content reconstruction loss $\mathcal L_{GAE^{content}}$ is relatively small.
However, we believe that content reconstruction loss is also indispensable.
The CAE module likes a residual network to supplement the high-quality content information to the GAE module layer-by-layer. Then, the content reconstruction loss ensures the middle layer representation learned by the GAE module contains more content information of the raw data.

Throughout these three ablation experiments, it turns out that each module improves the clustering performance from different aspects and is meaningful.

\section{Conclusion} \label{Sec:5}
We propose a cross-attention fusion based enhanced graph convolutional network for subspace clustering (CaEGCN), which connects the CAE and GAE modules layer-by-layer through the cross-attention fusion module, and strengthens the essential information.
The fusion representation is used as the input of the GAE module. 
The novel graph reconstruction loss and content reconstruction loss in the GAE module further ensure the middle layer representation more appropriate for clustering.
Finally, we build the self-supervised module to train the entire end-to-end model.
The excellent experimental results on various datasets prove the superiority of the proposed methods.

\section*{Acknowledgements}
The research project is partially supported by National Natural Science Foundation of China under Grant No. U19B2039, 61906011, 61632006, 61772048, 61672071, U1811463, 61806014, Beijing Natural Science Foundation No. 4204086, Beijing Municipal Science and Technology Project
No. KM202010005014, KM201910005028, Beijing Talents Project (2017A24), Beijing Outstanding Young Scientists Projects (BJJWZYJH01201910005018).

\bibliographystyle{IEEEtran}
\bibliography{reference}

\begin{IEEEbiography}[{\includegraphics[width=1in,height=1.25in,clip,keepaspectratio]{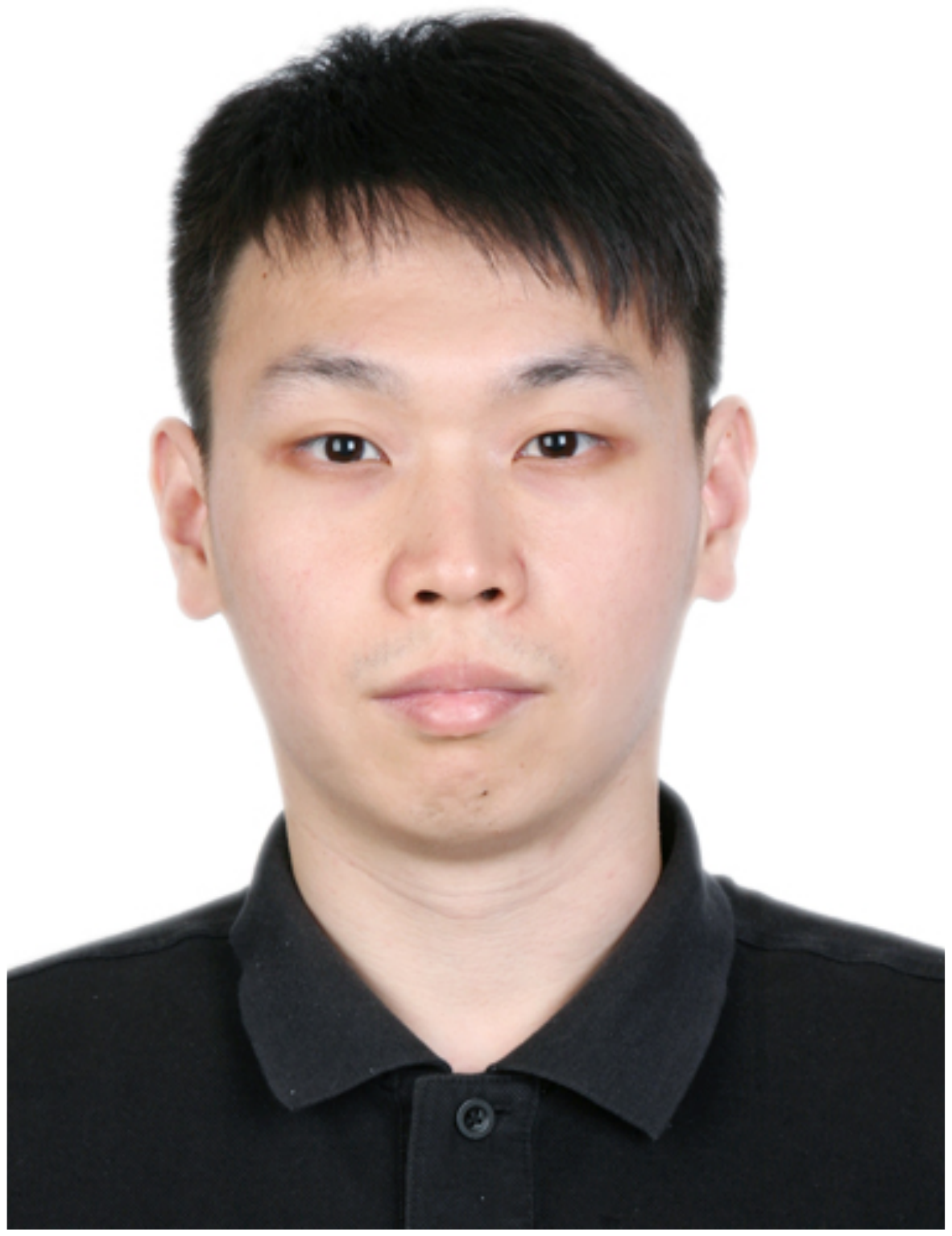}}]
{Guangyu Huo} received the B.Sc. degree in IoT Engineering and the M.S. degree in Computer Science from Beijing University of Technology, China in 2016 and 2019, where he is currently working toward the PhD. degree in Control Science and Engineering.
His current research interests include intelligent transportation, computer
vision, pattern recognition and deep learning.
\end{IEEEbiography}

\begin{IEEEbiography}[{\includegraphics[width=1in,height=1.25in,clip,keepaspectratio]{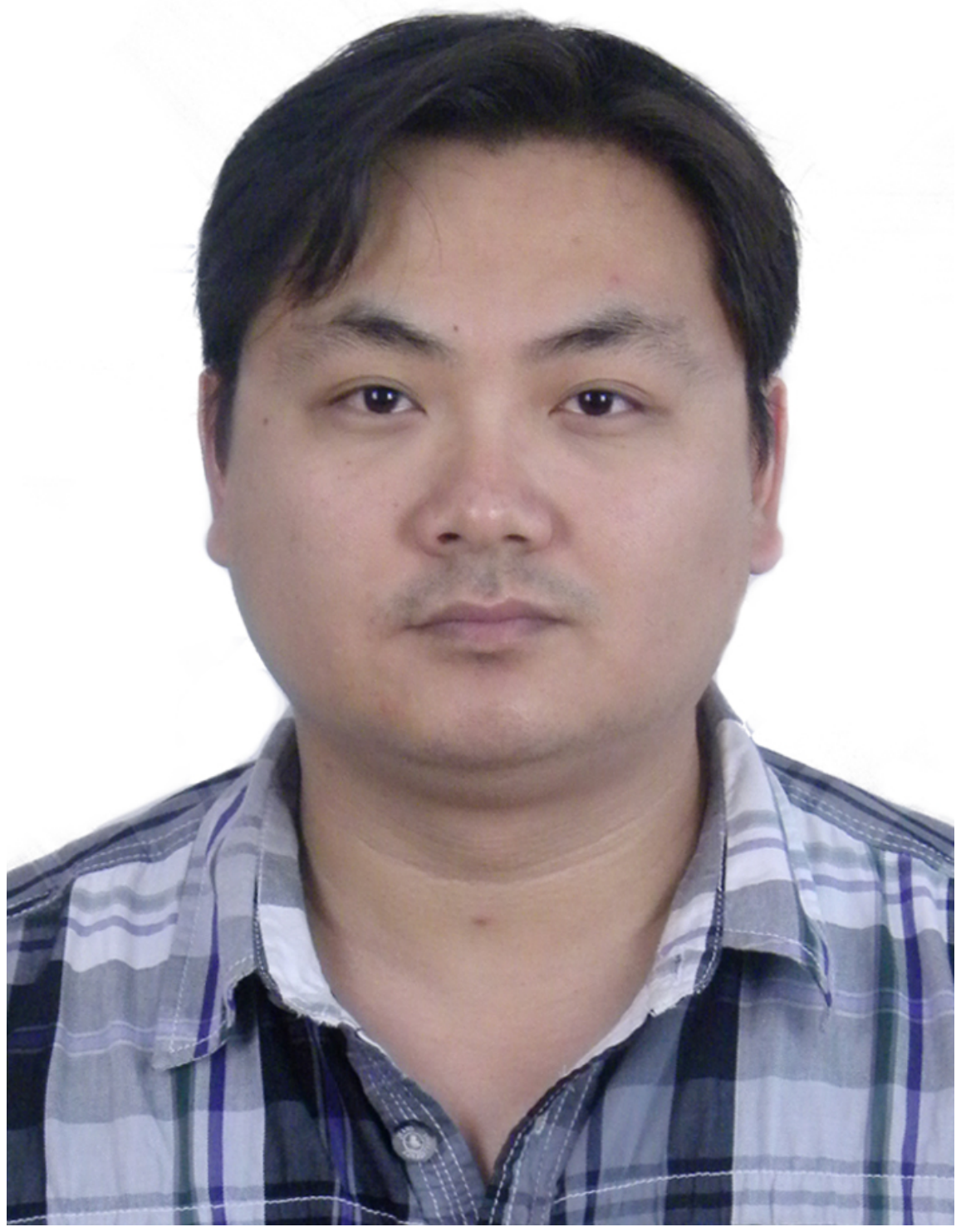}}]
{Yong Zhang} (M'12) received the Ph.D. degree in computer science from the BJUT, in 2010. He is currently an Associate Professor in computer science in BJUT. His research interests include intelligent transportation system, big data analysis and visualization, computer graphics.
\end{IEEEbiography}

\begin{IEEEbiography}[{\includegraphics[width=1in,height=1.25in,clip,keepaspectratio]{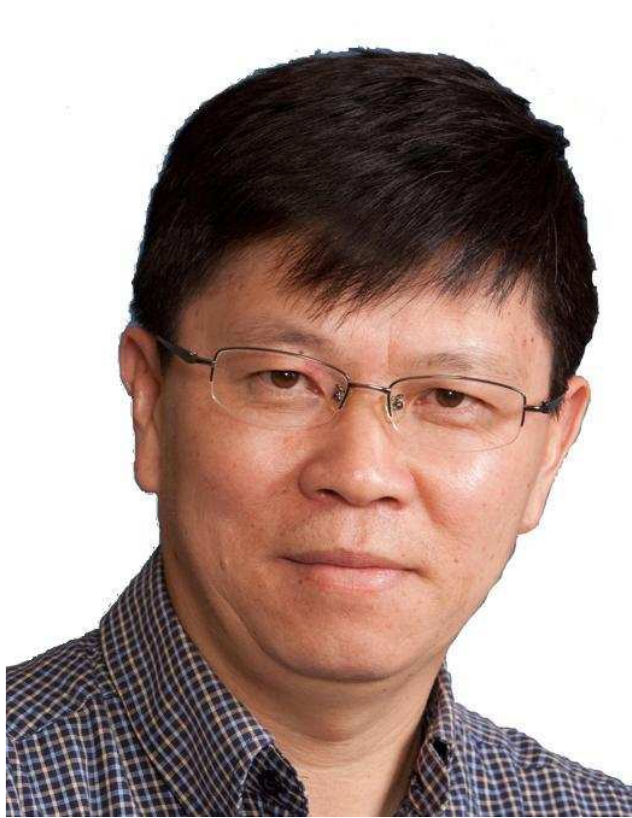}}]{Junbin Gao} graduated from Huazhong University of Science and Technology (HUST), China in 1982 with a BSc in Computational Mathematics and obtained his PhD from Dalian University of Technology, China in 1991. He is  Professor of Big Data Analytics in the University of Sydney Business School at the University of Sydney and was a Professor in Computer Science in the School of Computing and Mathematics at Charles Sturt University, Australia. He was a senior lecturer, a lecturer in Computer Science from 2001 to 2005 at the University of New England, Australia. From 1982 to 2001 he was an associate lecturer, lecturer, associate professor, and professor in Department of Mathematics at HUST. His main research interests include machine learning, data analytics, Bayesian learning and inference, and image analysis.
\end{IEEEbiography}

\begin{IEEEbiography}[{\includegraphics[width=1in,height=1.25in,clip,keepaspectratio]{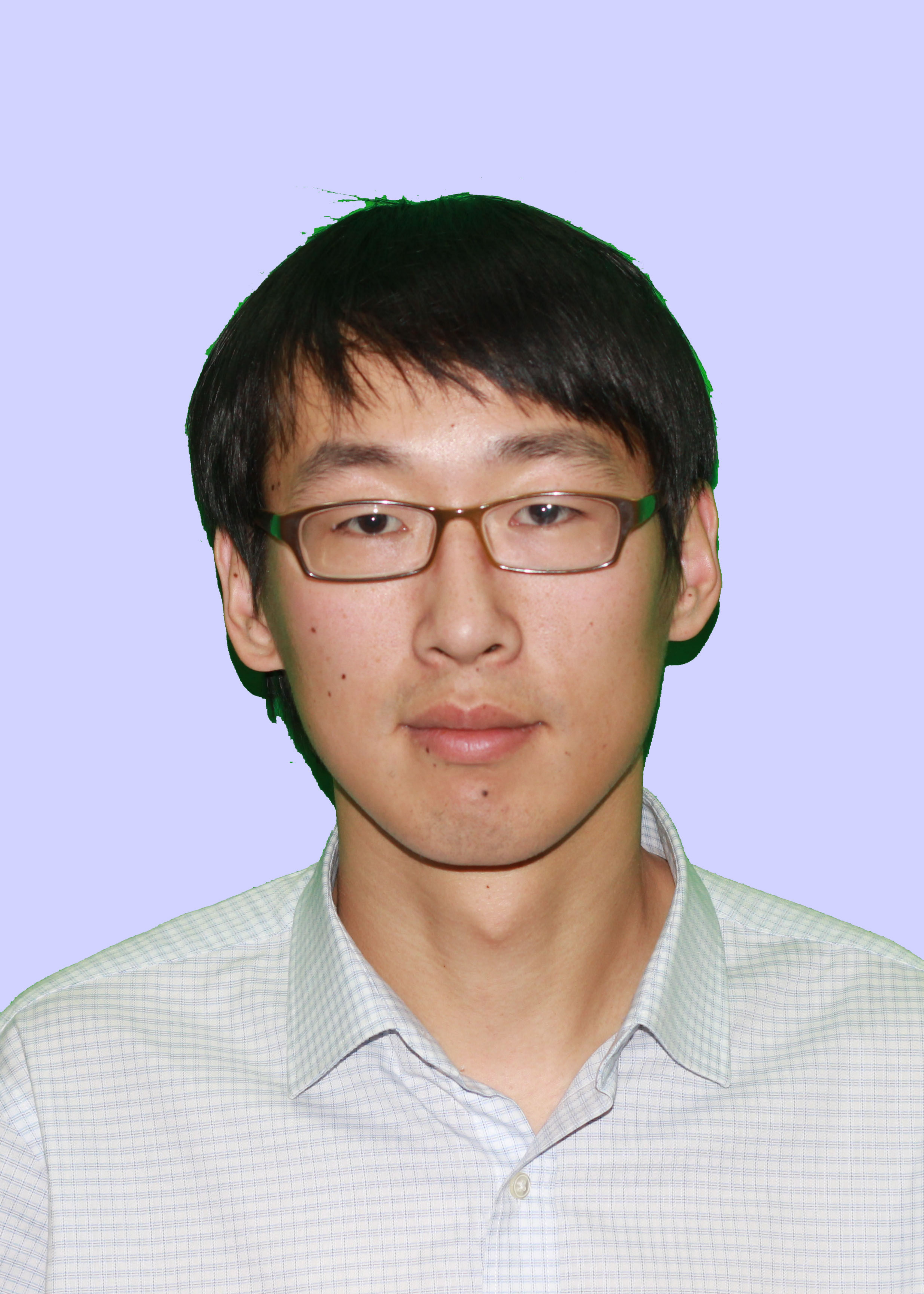}}]
{Boyue Wang} received the B.Sc. degree in Computer Science from Hebei University of Technology, China in 2012 and obtained PhD from Beijing University of Technology, China in 2018. He is a postdoctor in the Beijing Municipal Key Laboratory of Multimedia and Intelligent Software Technology,
Beijing University of Technology, Beijing.
His current research interests include computer
vision, pattern recognition, manifold learning and kernel methods.
\end{IEEEbiography}

\begin{IEEEbiography}[{\includegraphics[width=1in,height=1.25in,clip,keepaspectratio]{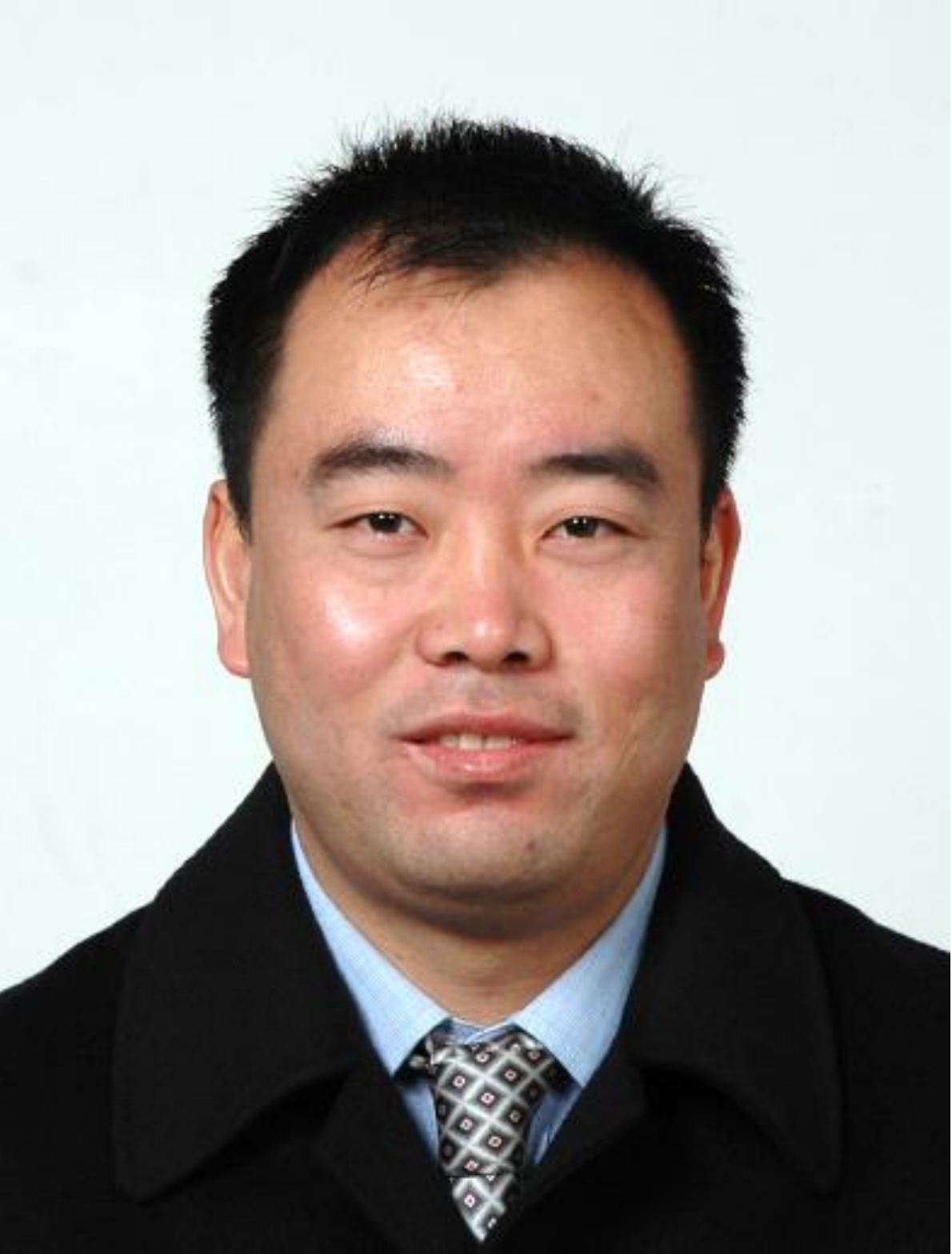}}]
{Yongli Hu} received his Ph.D. degree from Beijing University of Technology in 2005. He is a professor in the Faculty of Information Technology at Beijing University of Technology. He is
a researcher at the Beijing Municipal Key Laboratory of Multimedia and Intelligent Software Technology.
His research interests include computer graphics, pattern recognition and multimedia technology.
\end{IEEEbiography}

\begin{IEEEbiography}[{\includegraphics[width=1in,height=1.25in,clip,keepaspectratio]{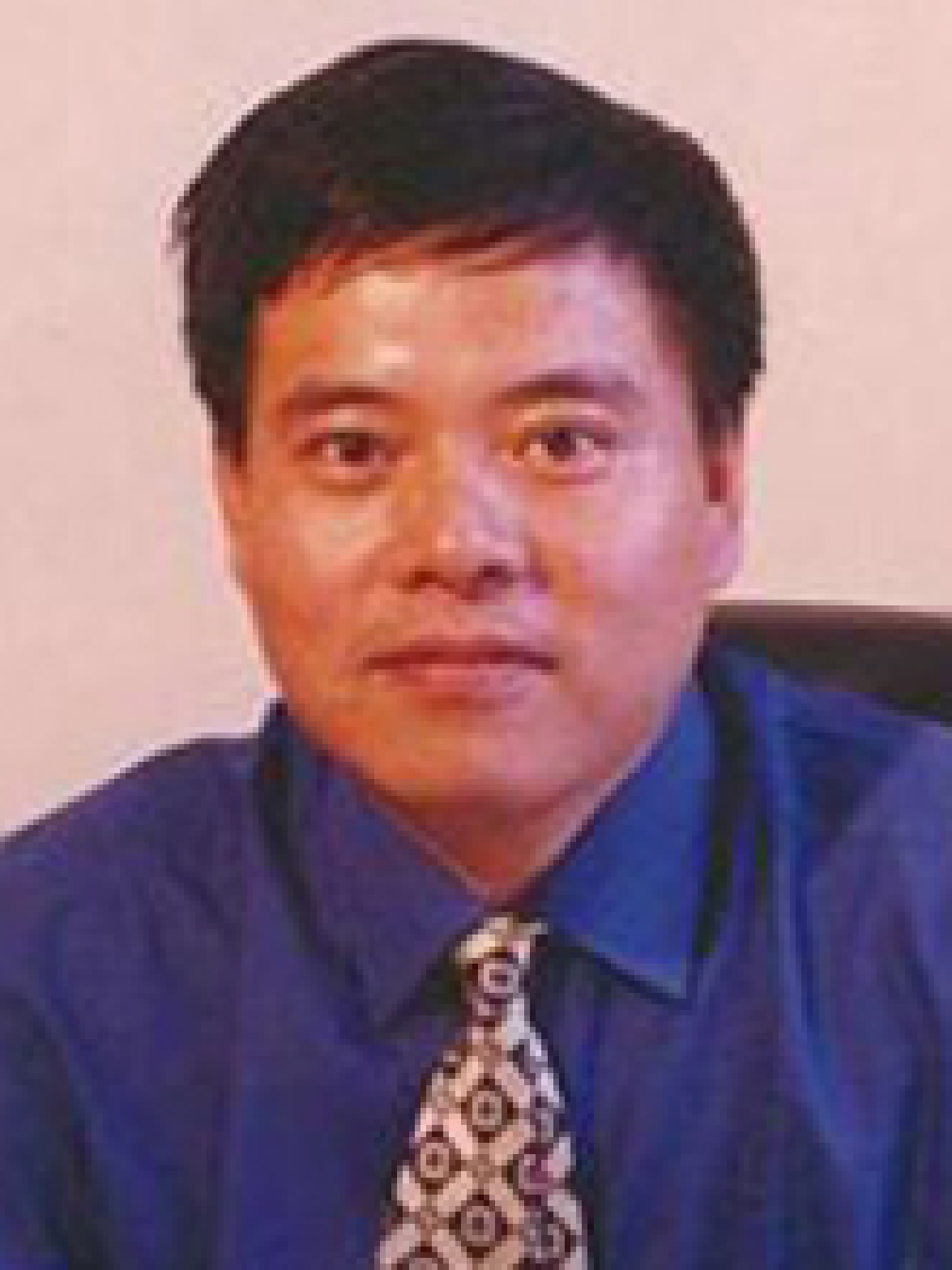}}]
{Baocai Yin} received his Ph.D. degree from Dalian University of Technology in 1993. He is a professor in the Faculty of Information Technology at Beijing University of Technology.
He is a researcher at the Beijing Municipal Key Laboratory of Multimedia and Intelligent Software Technology.
He is a member of China Computer Federation. His
research interests cover multimedia, multifunctional perception, virtual reality and computer graphics.
\end{IEEEbiography}

\end{document}